\documentclass[letterpaper, 10 pt, conference]{ieeeconf}  

\IEEEoverridecommandlockouts                              

\usepackage{amssymb, amsmath}
\usepackage{subcaption}
\usepackage{graphicx}
\usepackage{wrapfig}
\usepackage{bm}
\usepackage{hyperref}
\usepackage{cite}
\usepackage{etoolbox}

\usepackage[font=small,labelfont=bf]{caption}

\newcommand{\navigait}{\textsc{NaviGait}}

\newcommand{\R}{\mathbb{R}}

\title{\LARGE \bf
\textsc{NaviGait}: Navigating Dynamically Feasible Gait Libraries using \\ Deep Reinforcement Learning}

\newif\ifanonymous
\anonymoustrue   
\anonymousfalse

\newbool{codelinks}
\setbool{codelinks}{false}  

\ifanonymous
    \author{Anonymous Authors}
\else
    \author{Neil Janwani$^{\dagger}$, Varun Madabushi$^{\dagger}$, and Maegan Tucker
    \thanks{$\dagger$ These authors contributed equally to this work.}
    \thanks{This work is supported by the Georgia Tech Institute for Robotics and Intelligent Machines (IRIM) and NSF (CPS Award \#2440387)}
    \thanks{Authors are with the Dynamic Mobility Lab at Georgia Tech, Atlanta, U.S. \texttt{\{njanwani, vmadabushi3, mtucker34\}@gatech.edu}}
    }
\fi

\begin{document}

\maketitle
\thispagestyle{empty}
\pagestyle{plain}

\begin{abstract}

Reinforcement learning (RL) has emerged as a powerful method to learn robust control policies for bipedal locomotion.
Yet, it can be difficult to tune desired robot behaviors due to unintuitive and complex reward design. 
In comparison, trajectory optimization-based methods offer more tuneable, interpretable, and mathematically grounded motion plans for high-dimensional legged systems.
However, these methods often remain brittle to real-world disturbances like external perturbations.
In this work, we present \navigait, a hierarchical framework that combines the structure of trajectory optimization with the adaptability of RL for robust and intuitive locomotion control. 
\navigait~leverages RL to synthesize new motions by selecting, minimally morphing, and stabilizing gaits taken from an offline-generated gait library.
\navigait~results in walking policies that match the reference motion well while maintaining robustness comparable to other locomotion controllers.
Additionally, the structure imposed by \navigait~drastically simplifies the RL reward composition.
Our experimental results demonstrate that \navigait~enables faster training compared to conventional and imitation-based RL, and produces motions that remain closest to the original reference.
Overall, by decoupling high-level motion generation from low-level correction, \navigait~offers a more scalable and generalizable approach for achieving dynamic and robust locomotion. Videos and the full framework are publicly available at \href{https://dynamicmobility.github.io/navigait/}{\texttt{dynamicmobility.github.io/navigait}}.


\end{abstract}

\section{INTRODUCTION}

Dynamic legged locomotion in real-world environments demands both precise coordination and adaptive robustness.
Classical trajectory optimization methods such as Hybrid Zero Dynamics (HZD) \cite{westervelt2003hybrid, grizzle20103d} have long provided a principled approach to gait generation in high-dimensional legged systems with explicit impact dynamics.
These methods yield output reference trajectories with formal guarantees on stability and convergence, and have been successfully applied to a wide range of bipedal robots \cite{hereid2018dynamic,reher2019dynamic,tucker2023robust, ghansah2023humanoid}.
However, a well-known limitation of these approaches is their reliance on idealized models and assumptions, and that the trajectory generation is often too slow to run for online stabilization.
Thus, without online stabilization, the generated reference trajectories often fail to accommodate for real-world disturbances such as external perturbations or variations in terrain, which are inevitable in natural environments.

Meanwhile, reinforcement learning (RL) has emerged as a powerful alternative for synthesizing robust locomotion policies that learn through experience \cite{hwangbo2019learning, peng2017deeploco, rudin2022learning}.
RL policies can handle rich sensory feedback and learn to maintain stability across unstructured settings.
However, these capabilities come at the cost of significant sample complexity, long training times, and often opaque control strategies.
Furthermore, in locomotion settings, crafting reward functions that encourage natural, stable, and goal-directed behavior remains a nontrivial challenge \cite{lee2020learning, kumar2021rma}.
Due to this challenge, existing methodologies often result in policies that lack interpretability, transferability, and tunability.
Furthermore, RL alone does not easily incorporate prior knowledge about the structure and kinematics of natural-looking gaits. 

    \begin{figure}
        \centering
        \includegraphics[width=\linewidth]{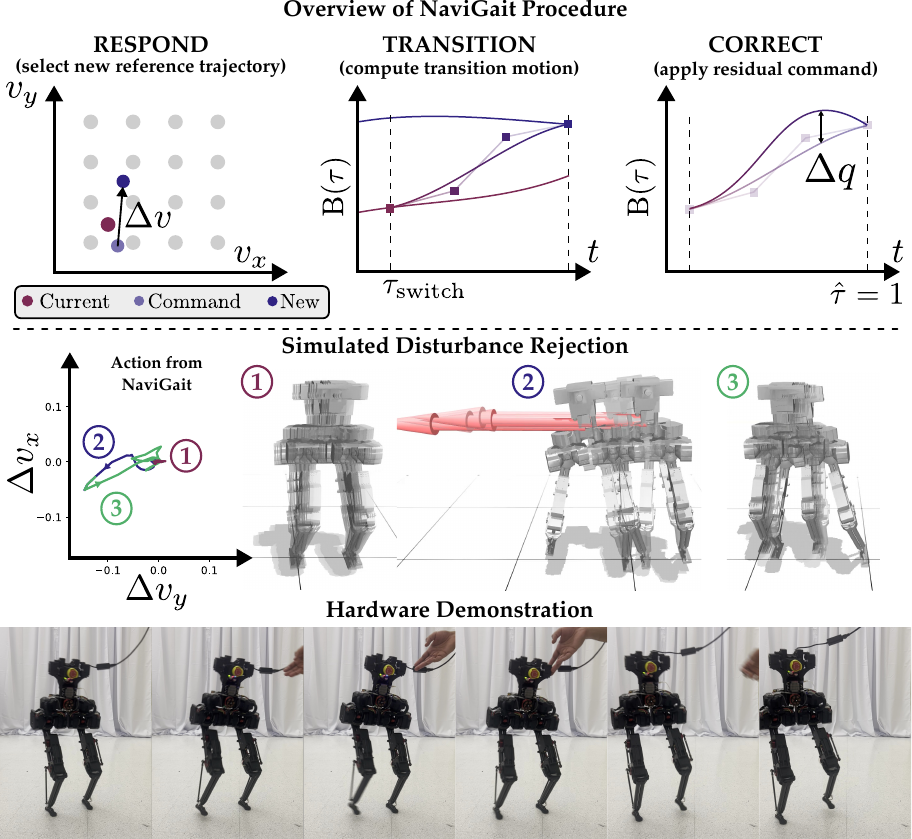}
        \caption{\navigait~simplifies reward design for bipedal locomotion by seamlessly integrating offline-generated gait libraries with RL.
        At every inference step, \navigait~(1) selects a reference trajectory, (2) smoothly transitions between the current motion and the newly selected reference motion, and (3) provides joint-level corrections for stabilization.
        We demonstrate \navigait~for velocity tracking and disturbance rejection tasks both in simulation and on hardware.
        } 
        \label{fig:hero}
        \vspace{-0.6cm}
    \end{figure}

In this paper, we propose to bridge the gap between these two paradigms by learning to smoothly warp trajectories from precomputed \textit{gait libraries}--collections of optimized physics-informed reference gaits spanning a range of velocities.
Our approach, denoted \navigait, learns a residual policy capable of transitioning between reference motions in the gait library while providing minimal joint-level corrections.
This residual formulation enables the RL policy to focus on stabilization and adaptation, while the underlying precomputed trajectory optimization handles the bulk of the motion synthesis.
As a result, \navigait~not only inherits the interpretability and structure of optimized gaits but also gains the robustness and flexibility of learned control.
    

    A key advantage of our approach is that it simplifies the reward design.
    Since gait libraries already encode how motions should look and feel, a learned policy does not need to ``rediscover'' good locomotion from scratch as it does in the conventional case.
    Instead, \navigait~learns relatively low-dimensional corrections that align behavior with task goals while remaining grounded in physically meaningful references. This results in two additional advantages: a significantly reduced training time and resulting policies that remain close to the original reference motion. 

    We validate \navigait~on Bruce \cite{Liu2022bruce}, a low-cost humanoid platform, across core facets of learned locomotion--training speed, naturalness, high-level velocity tracking and disturbance rejection.
    Our results demonstrate that \navigait~is capable of maintaining similar or better performance to other state-of-the-art RL methods, while being faster to train and easier to shape behaviors to match a desired style.
    
    Our specific contributions are as follows:

    \begin{enumerate}
        \item We present \navigait, a novel hierarchical framework that integrates a gait library of optimized physics-informed reference motions with a residual RL policy capable of continuously modulating between them.
        \item We open-source the first, to our knowledge, JaX-compatible  \cite{jax2018github} implementation of smooth continuous gait reference interpolation and blending, enabling just-in-time compilation and easy parallelization.
        \item We demonstrate that \navigait~simplifies reward design, accelerates training, and has improved imitation accuracy compared to two baselines: canonical RL without any motion references \cite{zakka2025mujoco}, and imitation learning with gait library references \cite{li2021reinforcement}. 
        \item We further motivate our residual-based architecture by learning two stylistically different locomotion policies without altering the controller structure.
        \item We validate \navigait~on the BRUCE humanoid both in simulation and on hardware, achieving robust stabilization in the presence of external perturbations.
    \end{enumerate}
    
    Ultimately, by combining the strengths of model-based motion planning and reinforcement learning, \navigait~offers a scalable and generalizable solution for achieving efficient and natural dynamic locomotion in the real world.

\section{RELATED WORK}

Our work is motivated by the observation that gait libraries contain a wealth of offline generated reference motions with local stability guarantees.
Importantly, these libraries are relatively fast to produce and can be easily tuned to new robots or to robots with changing model parameters (e.g. wearable robots).
However, the limitation with this approach is the reliance on heuristic hand-tuned regulators to track time-varying velocity commands and reject disturbances.
In our work, we seek to replace the heuristic and brittle nature of gait library regulators with a learned residual control policy.
We hypothesize that our approach will improve sample efficiency compared to RL trained without reference motions while maintaining base velocity tracking and disturbance rejection capabilities.
Details on existing methods for gait libraries and reinforcement learning towards obtaining stable bipedal locomotion are provided below.

\subsection{Gait Generation}
Bipedal robot walking is a problem long-studied by control theorists and roboticists, due to its hybrid and (in some cases) underactuated structure.
One approach that has historically seen success is the Hybrid Zero Dynamics (HZD) method \cite{westervelt2003hybrid} which generates periodic reference motions with local stability guarantees. The HZD method represents the robot dynamics as a continuous domain (corresponding to the single stance phase) connected with a reset map capturing the impact dynamics.
Under these dynamics, a nonlinear program (NLP) can be solved to generate periodic trajectories in the robot's joint space that correspond to walking behavior. These behaviors are physically realized through joint-space tracking (PD control), and by construction lead to walking that is provably stable under bounded disturbance \cite{tucker2023input}.
Gait properties can be easily specified by shaping constraints and costs and solving the corresponding NLP.

A drawback of this method is that generation of a single gait is too time-consuming for online dynamic re-planning.
Previous works have addressed this shortcoming through ``gait libraries,'' or sets of reference trajectories, and transition behaviors which smoothly switch between gaits \cite{westervelt2018feedback, nguyen2020dynamic}.
However, the stabilizing behaviors must be heuristically designed and tuned, and can only marginally deviate from the nominal trajectory.
Overall, this method has achieved dynamic, stable, and visually appealing walking, but lacks robustness due to its limited disturbance rejection ability.

\subsection{Reinforcement Learning}
Reinforcement Learning (RL) is another popular approach to stabilization. In RL, behavior is specified through a scalar reward function which the learning algorithm seeks to maximize over time.
However, walking is a complex, multifaceted behavior, and simple costs (e.g. ``go forward'') tend to result in behaviors which may not be safe or visually appealing \cite{siekmann2021sim}.
Thus, special care must be taken to ensure that the reward structure not only completely specifies the set of desired behaviors, but also guides the learning process towards desirable solutions.
When paired with long training time, reward design can be a taxing and unintuitive endeavor.

For RL applied to bipedal locomotion, it is common practice to introduce imitation rewards, where the policy is encouraged to replicate behaviors from a library of reference motions \cite{peng2018deepmimic, hasenclever2020comic, li2021reinforcement, escontrela2022adversarial, yu2022dynamic, he2025hoverversatileneuralwholebody, grandia2025design}.
Imitating a dataset of reference human motions naturally results in robot behaviors that are more human-like, while avoiding the need to specify and tune a potentially complex reward function.
The main challenge remains the curation of the reference motion dataset, which must be collected from motion-capture recordings of human behaviors or through character animations, and subsequently re-targeted onto the robot embodiment of choice.

\subsection{Reinforcement Learning with Offline Generated Gaits}
Notably, there exists prior work in RL that takes advantage of offline-generated reference motions. Within this class of approaches, there are still differences in how to integrate the motions into the learning process.
One approach is to apply the reference trajectory as a fixed input to a policy network which then produces online corrections to the reference\cite{xie2018feedback}.
This strategy blends open-loop reference tracking with learned closed-loop adjustments but is limited by its reliance on a single reference motion, which may restrict generality across different terrain or tasks.

In \cite{li2021reinforcement}, a gait library is used to store a series of gaits corresponding to a wide range of commands.
When the operator commands a particular velocity, the nearest gait corresponding to that behavior is supplied to the policy network in a manner similar to \cite{xie2018feedback}.
This work leverages a wide variety of reference motions to generate a diverse set of behaviors.
However, the selection of architecture in this work permits the network to output policies that are drastically different from the reference motion set, which can be a drawback if style and aesthetics are a priority.


In comparison to these existing methodologies, 
\navigait~provides a unique architecture in its ability to both choose new reference motions and provide corrections through simultaneous residual joint commands and changes to the reference velocity.
As in \cite{li2021reinforcement}, we use a pre-generated gait library computed via trajectory optimization.
However, our library is a smooth and continuous collection of gaits rather than a grid as in \cite{li2021reinforcement}, promoting smooth transitions to new desired motions.
Moreover, \navigait~also features residual control as in \cite{xie2018feedback} to ensure that the resulting joint trajectories remain near to those of the gait library.
\navigait~circumvents the weakness of residual control, namely its lack of disturbance rejection, by altering the reference velocity as described above.
Thus, our approach strategically employs reinforcement learning to stabilize the robot while maintaining the look and feel of the gait library.

\section{Method}
The \navigait~framework has two major components: 1) smooth interpolation across a gait library to provide a continuous space of reference motions; 2) a deep reinforcement learning policy to provide residual changes to the target joint angles and to the high-level velocity commands.

\subsection{Robot Model}

We choose to model the bipedal robot using the \textit{floating-base model}, which  appends the position and orientation of the robot's base frame to the local joint configuration, providing important global information.
The tradeoff is that ground contacts must be enforced explicitly as holonomic constraints at the feet, rather than being implicitly assumed as in a \textit{pinned-base model}.

To represent the floating-base, let $R_b$ be the body reference frame rigidly attached to the base frame of the robot.
This frame is relative to the fixed inertial frame attached to the world, denoted as $R_0$.
The \textit{extended coordinates} $q_e$ of the robot are defined as $q_e := (p_b^T, \phi_b^T, q_l)^T \in \R^3 \times SO(3) \times \mathcal{Q}_l$ with $p_b \in \R^3$ and $\phi_b \in SO(3)$ being the Cartesian position and orientation, respectively, of $R_b$ relative to $R_0$, and $q_l \in \mathcal{Q}_l \subset \R^{n_l}$ being the local coordinates which are the joint-level positions of the robot ($n_l = 14$ for BRUCE).
The entire configuration space of the extended coordinates is denoted by $q_e \in \mathcal{Q}$ with the full system state belonging to the tangent space, $x_e = (q_e^T, \dot{q}_e^T)^T \in \mathsf{T}\mathcal{Q}$.
For notation purposes, let $n_e \in \R_+$ be the total number of states within the extended configuration $q_e$.
In our model of BRUCE, $n_e = 41$.

\begin{figure}[tb]
    \centering
    \begin{subfigure}[b]{0.15\textwidth}
        \centering
        \includegraphics[width=\textwidth]{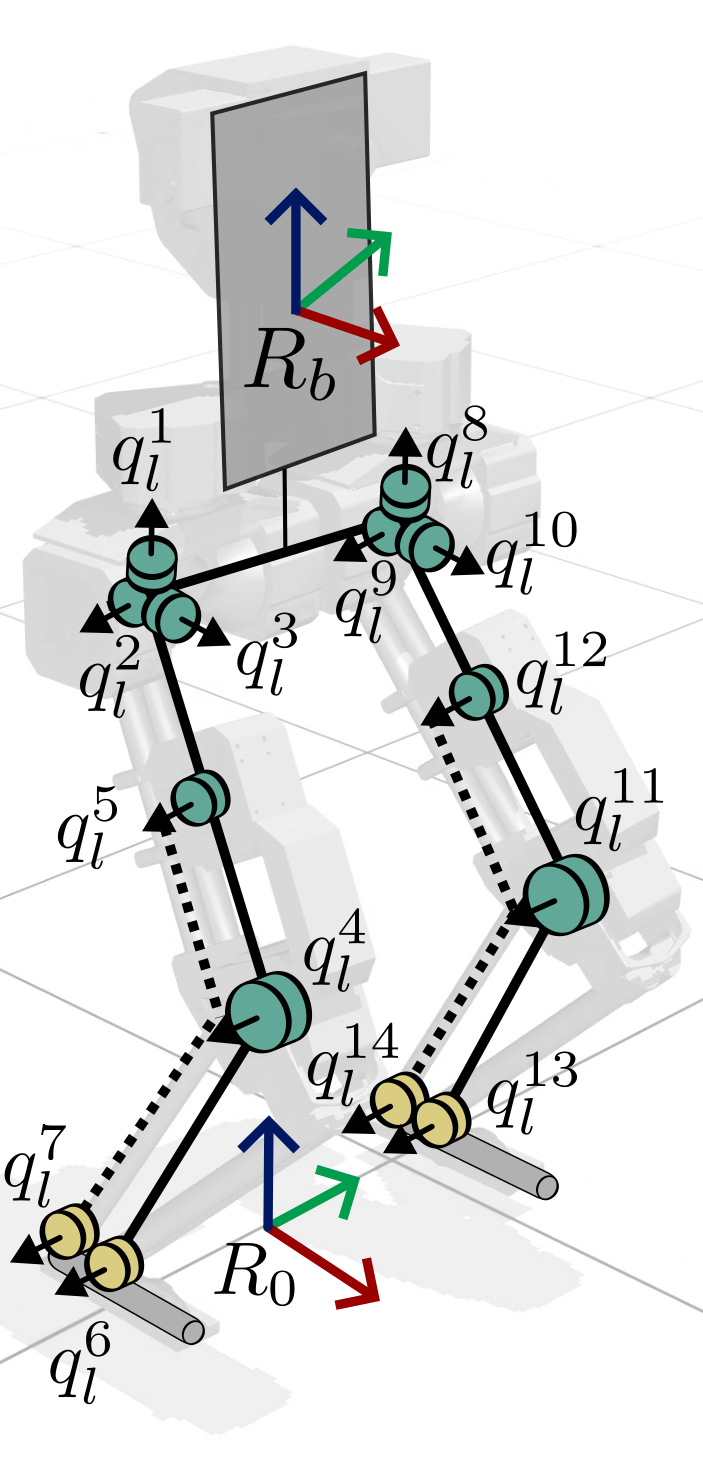}
    \end{subfigure}
    \begin{subfigure}[b]{0.15\textwidth}
        \centering
        \includegraphics[width=\textwidth]{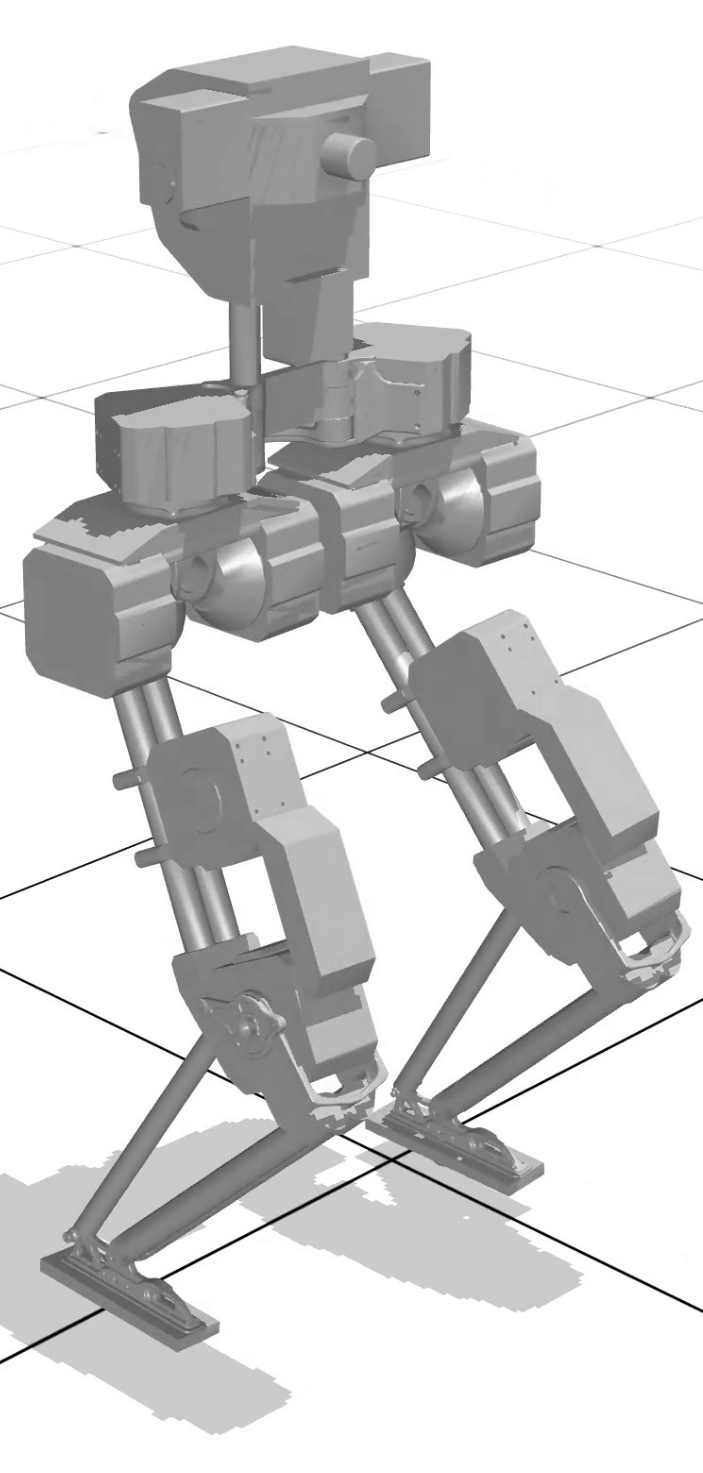}
    \end{subfigure}
    \begin{subfigure}[b]{0.14\textwidth}
        \centering
        \includegraphics[width=\textwidth]{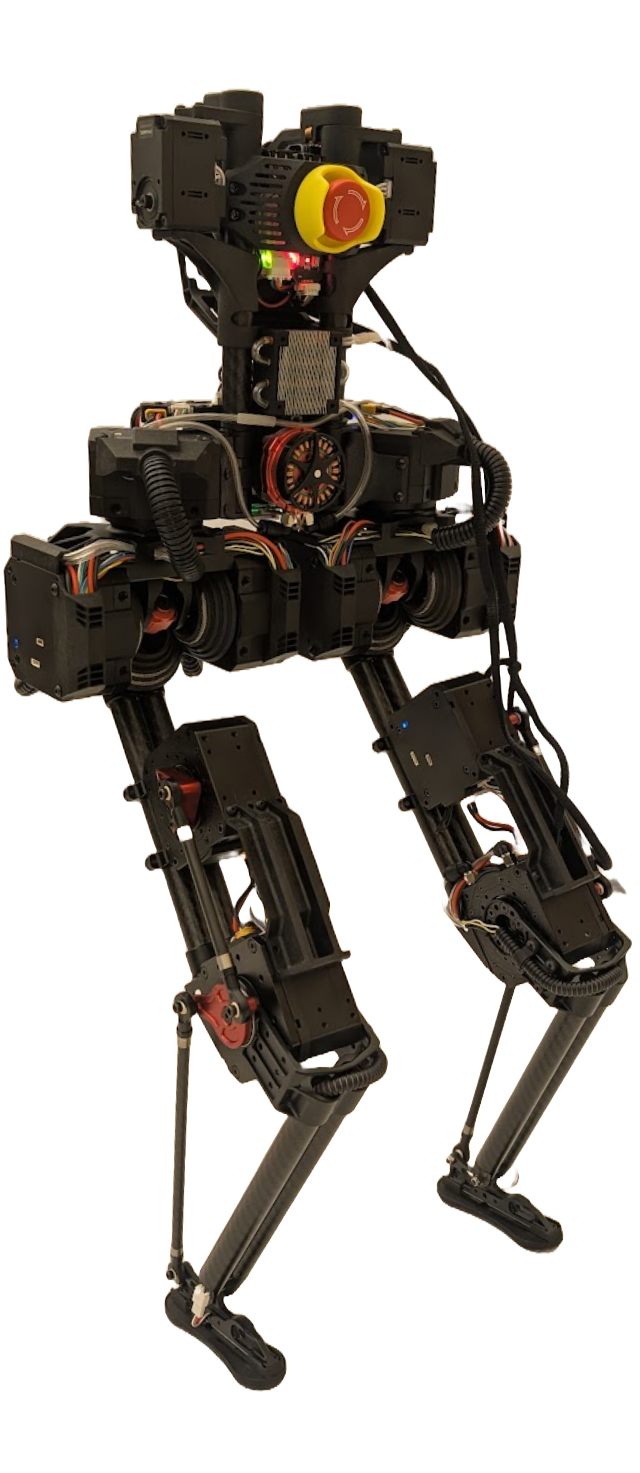} 
    \end{subfigure}
    \caption{Illustration of the BRUCE kinematic model, its simulation, and the robot hardware. }
    \label{fig:bruce_structure}
    \vspace{-5mm}
\end{figure}

It is worth noting that we choose to model the feet of BRUCE as a line-contact (due to small foot width and no ankle roll mechanism).
Moreover, we enforce the 4-bar linkage mechanism as an additional 4-dimensional holonomic constraint.
This results in a system with one degree of under-actuation.
Figure \ref{fig:bruce_structure} depicts the structure of the robot model.
The yellow color denotes the passive joints, while the blue color denotes the actively controlled joints.
The constraint enforced by the linkage bar is depicted with the dotted line.


\begin{figure*}[tb]
\vspace{4mm}
    \centering
    \includegraphics[width=0.8\linewidth]{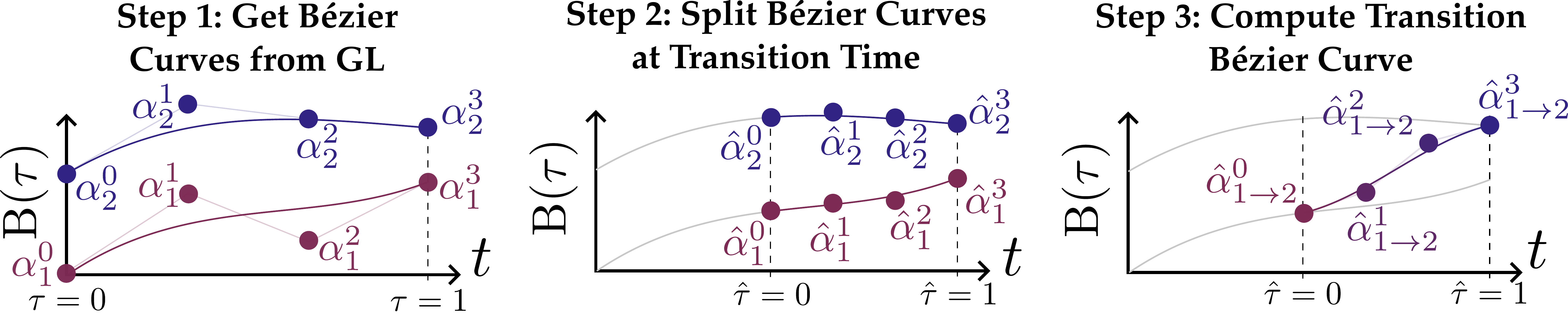}
    \caption{The diagram illustrates the procedure for smoothly interpolating between different gait references. Importantly, the implementation allows for continuous reference tracking by recursively repeating the illustrated process for some new phasing parameterization $\hat{\tau} \in [0,1]$.}
    \label{fig:bezier}
    \vspace{-4mm}
\end{figure*}    

\subsection{Gait Reference Generation Procedure}
\label{sec:GaitGen}
We generate reference motions\ifbool{codelinks}{\footnote{\label{fn:trajopt}Gait Gen. Repo: \href{https://anonymous.4open.science/r/bruce_trajopt-7DDF/}{https://anonymous.4open.science/r/bruce\_trajopt-7DDF/}}
}{}using the FROST package \cite{hereid2017frost} which transcribes a \textit{Hybrid Zero Dynamics} optimization problem into a direct collocation trajectory optimization problem.
The optimization problem is set up to solve for a set of B\'ezier curve parameters $\alpha = \R^{n_o \times b+1}$ that parameterize the desired outputs (of dimension $n_o$), with $b \in \mathbb{Z}_+$ denoting the degree of the corresponding curves.
For BRUCE, there are 10 outputs ($n_o = 10$), which are selected to be the positions of the actuated joints. The B\'ezier curves are of order $b=7$.
It is standard to write the full NLP as:

\small{
\begin{align}
    X(\alpha)^* = ~&\underset{X(\alpha)}{\textrm{argmin}} ~\mathcal{J}(X(\alpha)) \notag \\
   \textrm{s.t.} & \quad \textrm{Closed Loop Dynamics} 
   \label{eqn:dynamics}\\
    & \quad \textrm{Impact Invariance} 
    \label{eqn:invariance}\\
    & \quad \textrm{Physical Feasibility} 
    \label{eqn:feasibility}\\
    & \quad \textrm{Additional Constraints}
    \label{eqn:constraints}
\end{align}
}
\normalsize
with, $X(\alpha)$ denoting the joint trajectories parameterized by $\alpha$, and $\mathcal{J}$ denoting the cost function.
Our cost function encourages minimizing the overall torque input, and encourages base velocity tracking.
\ifbool{codelinks}{The full formulation can be found in the gait generation repository$^{\ref{fn:trajopt}}$.}{}

The purpose of each of the constraints is as follows: \eqref{eqn:dynamics} and \eqref{eqn:invariance} ensure that the trajectories are feasible with respect to the continuous swing-phase dynamics and the discrete impact dynamics; \eqref{eqn:feasibility} enforces joint limits in position, velocity, and torque; and \eqref{eqn:constraints} encodes gait characteristics such as step clearance, step length, and timing.

Importantly, the cost function and additional constraints (4) are often used to shape the desired behavior of the resulting gait patterns.
This has been demonstrated in earlier work to be a straightforward method of personalizing and ``stylizing'' the desired robotic motions \cite{tucker2021preference}.
Solving the NLP in the above formulation is relatively fast (e.g. under 30 seconds on an Apple M3 Pro), allowing for quicker turnaround times when tuning gait constraints compared to tuning reward weights in an RL context.
For this reason, a key advantage of our work is the ability to systematically combine the ease of tuning afforded via offline gait libraries with the robustness afforded via reinforcement learning.

\subsection{Gait Library Interpolation}
\label{sec:gaitlibrary}
In this work, we choose to represent the gait library as a continuous space of reference motions. This decouples the gait library structure from the policy structure (i.e., the discretization of the gait library is not directly coupled to the corrections the network has to make). 
To generate a continuous space from the discrete set of gaits generated in section \ref{sec:GaitGen}, we leverage the properties of B\'ezier polynomials: namely their ability to be interpolated and blended using only information of the control points. 
B\'ezier polynomials enable fast evaluation, curve splitting, and convex combinations all via matrix multiplications applied to the matrix of B\'ezier coefficients. The result is a computationally fast method of computing reference motions that smoothly connect the current reference gait to a gait corresponding to any desired velocity command $\hat{v} \in \R^2$. Moreover, these smooth transitions capture the look and feel of the gait library.

Notably the computational simplicity is especially important for ensuring JaX-compatibility. The JaX \cite{jax2018github} library permits Just-In-Time compilation and parallelization of Python code, provided that the code is comprised of compatible operations.
Importantly, our work provides the first JaX-compatible implementation of gait library operations, including B\'ezier curve blending, that makes it practical for parallel simulation, and thus simulation-based learning methods.

In the remainder of this section, we detail our procedure for generating smooth B\'ezier polynomial transitions from some gait with velocity $v_1$ to another with velocity $v_2$.
For a complete introduction to B\'ezier curves we refer the reader to \cite{pomaxBezier}. 
We will denote the time at the start of the transition as $t_1$, corresponding to a phasing variable of $\tau(t_1)$ with $\tau(t) := \frac{t-t_0}{T}$, $t_0$ being the time at the start of the stride and $T$ being the desired step duration. 

First, if $v_2$ does not correspond to an explicitly generated gait in the gait library, the desired reference gait is computed as the convex combination of the B\'ezier coefficients of the three nearest gaits within the library.
Then, the current spline and the desired spline are spliced at the phasing variable $\tau(t_1)$, resulting in two new splines $\hat{\alpha}_1 \in \R^{n_o \times b+1}$ and $\hat{\alpha}_2 \in \R^{n_e \times b+1}$ which are evaluated using the new phasing variable $\hat{\tau} := \frac{t-t_1}{T - (t_1 - t_0)}$ which maps the interval $[t_1, t_0 + T] \to [0, 1]$. The splicing operation is conveniently performed using a matrix multiplication operation as explained in Section 11 of \cite{pomaxBezier}.
Finally, a smooth transition gait can be obtained between the two new splines as:

\small
\begin{align}
    \hat{\alpha}_{1 \to 2} = \Big[
        &\hat{\alpha}_1^0 \quad \hat{\alpha}_1^1  \quad \hat{\alpha}_1^2 \quad \frac{\hat{\alpha}_1^3 + \hat{\alpha}_2^3}{2} \quad \frac{\hat{\alpha}_1^{b-3} + \hat{\alpha}_2^{b-3}}{2} \\
        & \dots \quad \hat{\alpha}_2^{b-2} \quad \hat{\alpha}_2^{b-1} \quad \hat{\alpha}_2^{b}
    \Big] \in \R^{n_e \times b + 1} \notag
\end{align}
\normalsize

where $\hat{\alpha}_i^j$ represents the $j$-th coefficient of the $i$-th gait and $b$ is the degree of the B\'ezier polynomial.
Figure \ref{fig:bezier} illustrates this smooth curve transition technique. 

\subsection{\navigait~Policy}
The \navigait~policy is a neural network which both chooses a reference trajectory and applies stabilizing control.
The policy network receives a user command, $v^d \in \R^2$, and a history of observations from the environment, and outputs both a residual for each actuated joint ($\Delta q \in \R^{n_o}$) and a reference velocity residual ($\Delta v = (\Delta v_x$, $\Delta v_y)$).
The velocity residuals are added to the user's requested velocity to form the velocity target, $\hat{v}^d = v^d + \Delta v$.
This velocity target is then input into the gait library to obtain the desired joint position $q$.
The joint targets are added to the joint residuals to form the motor position targets, $\hat{q}^d = q^d + \Delta q$ which are tracked using a joint-level PD controller running at 2000~Hz.
Lastly, histories of sensor data, desired joint and body states from the gait library, positions and velocities of the base relative to the stance foot, and the network's previous actions are fed back to the policy network.

\begin{figure}[t]
\vspace{2mm}
    \centering
    \includegraphics[width=\linewidth]{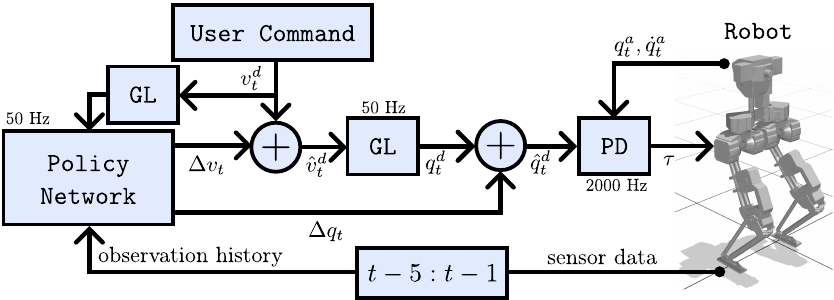} 
    \caption{Architecture of the \navigait~framework.}
    \label{fig:framework}
    \vspace{-4mm}
\end{figure}

The full policy implementation, including training, gait library generation, pre-generated gait libraries, and evaluation is provided in an open-source repository\footnote{\navigait~Repo: \href{https://github.com/dynamicmobility/navigait}{https://github.com/dynamicmobility/navigait}\label{navigait-code}}.
A diagram of the policy architecture at runtime is also provided in Fig. \ref{fig:framework}.

One note about \navigait~ is that despite the gait library only containing velocities in the forward and lateral directions (no turning gaits), \navigait~ is still capable of achieving turning by simply tracking a reference gait that is rotated with respect to the robot's forward direction. 

\noindent

\subsection{Learning Setup}
The \navigait~policy was trained using proximal policy optimization (PPO), implemented in \texttt{Brax} \cite{brax2021github}.

The observation for the policy network first consists of histories: (1) robot sensor data (gyroscope, accelerometer, base attitude, joint positions), (2) base and joint reference trajectories, and (3) previous network outputs.
We also provide the user velocity command, and the current reference joint and base velocities.
The value network receives the same inputs as the policy network, and additionally receives a privileged history of the cartesian base position and the currently applied disturbance force, which helps guide the learning process.

The reward structure is simple and consists of tracking, energy minimization, and smoothing terms.
The first three terms reward tracking the reference trajectories, the fourth term rewards minimizing the applied torque, and the last three encourage smoothness by discouraging the policy from rapidly changing its residual outputs. 

\small
\begin{align}
    R_{\navigait}^t(s, a) =~&R_{\text{Gait}}^t + R_{\text{Base Cartesian}}^t + R_{\text{Base Orientation}}^t + \\
    &R_{\text{torque}} + R_{\Delta v \text{ size}}^t + R_{\Delta v \text{ rate}} + R_{\Delta q \text{ rate}}. \notag 
\end{align}
\normalsize
In order to facilitate sim-to-real transfer, we implement domain randomization over the ground friction coefficient, damping, armature, joint friction, mass and center-of-mass of the torso, center-of-mass of links, PD gains of the joints, observation delay, robot starting position, and perturbations to the robot base frame.

The robot environment is implemented in MuJoCo JaX (MJX) as outlined in MuJoCo Playground \cite{zakka2025mujoco}.
We simulate the 4-bar linkage controlling BRUCE's ankle rotation through MuJoCo's soft equality constraint solver.
We observe that without the 4-bar mechanism in the model, the kinematic relationships are unchanged but the dynamics of the robot differ, which limits the policy sim-to-real transfer.
This finding is corroborated in \cite{tanaka2025MechanicalRL}.


\subsection{Hardware Implementation}
We evaluate our approach on the BRUCE robot \cite{Liu2022bruce} as shown earlier in Fig.  \ref{fig:bruce_structure}.
This robot is equipped with 5 proprioceptive actuators driving each leg, an ISM330DHCX inertial measurement unit (IMU), and a Khadas Edge 2 single-board computer (SBC).
A complementary filter runs on the SBC and produces estimates of the body attitude at 500~Hz.
The estimated body attitude quaternion along with raw measurements of the body rotation rate, body acceleration, joint position, and joint velocity, are serialized as Protobuf \cite{protobuf} messages and sent to an offboard computer via an ethernet connection.
The neural network controller runs on the offboard computer and is implemented in a client-server architecture.
The robot is the client, and periodically sends a Protobuf message containing all sensor measurements to the offboard computer.
The offboard computer, acting as the server, receives the sensor measurements and computes the joint targets via the \navigait~policy, which are sent back in turn as another Protobuf message.
The estimator and client, which run on the SBC, are written in a multithreaded fashion to ensure control commands and sensor measurements are applied at precisely 50~Hz with minimal delay.

\begin{figure*}[tb]
    \centering
    \includegraphics[width=\linewidth]{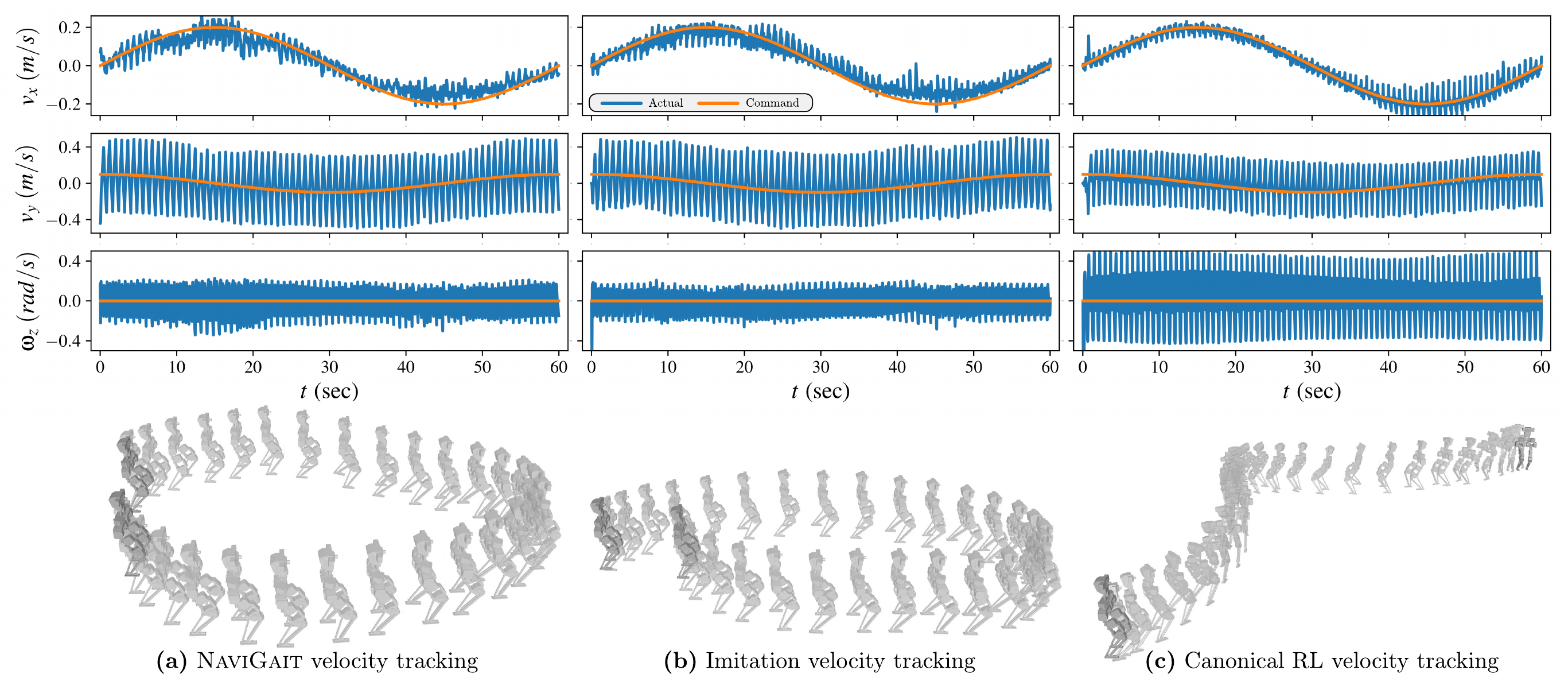}
    \caption{All three policies are shown tracking a time-varying ovular velocity command with zero angular velocity. Note that lateral robot velocity ($v_y$) is expected to have large deviations since BRUCE has underactuated dynamics in the frontal plane. An interesting observation is that \navigait~and Imitation RL exhibit better angular velocity tracking, and thus less overall drift, compared to Canonical RL.}
    \label{fig:velocity-tracking}
    \vspace{-4mm}
\end{figure*}

\begin{figure*}[tb]
    \centering
    \begin{subfigure}[htbp]{0.3\textwidth}
        \centering
        \includegraphics[width=\linewidth]{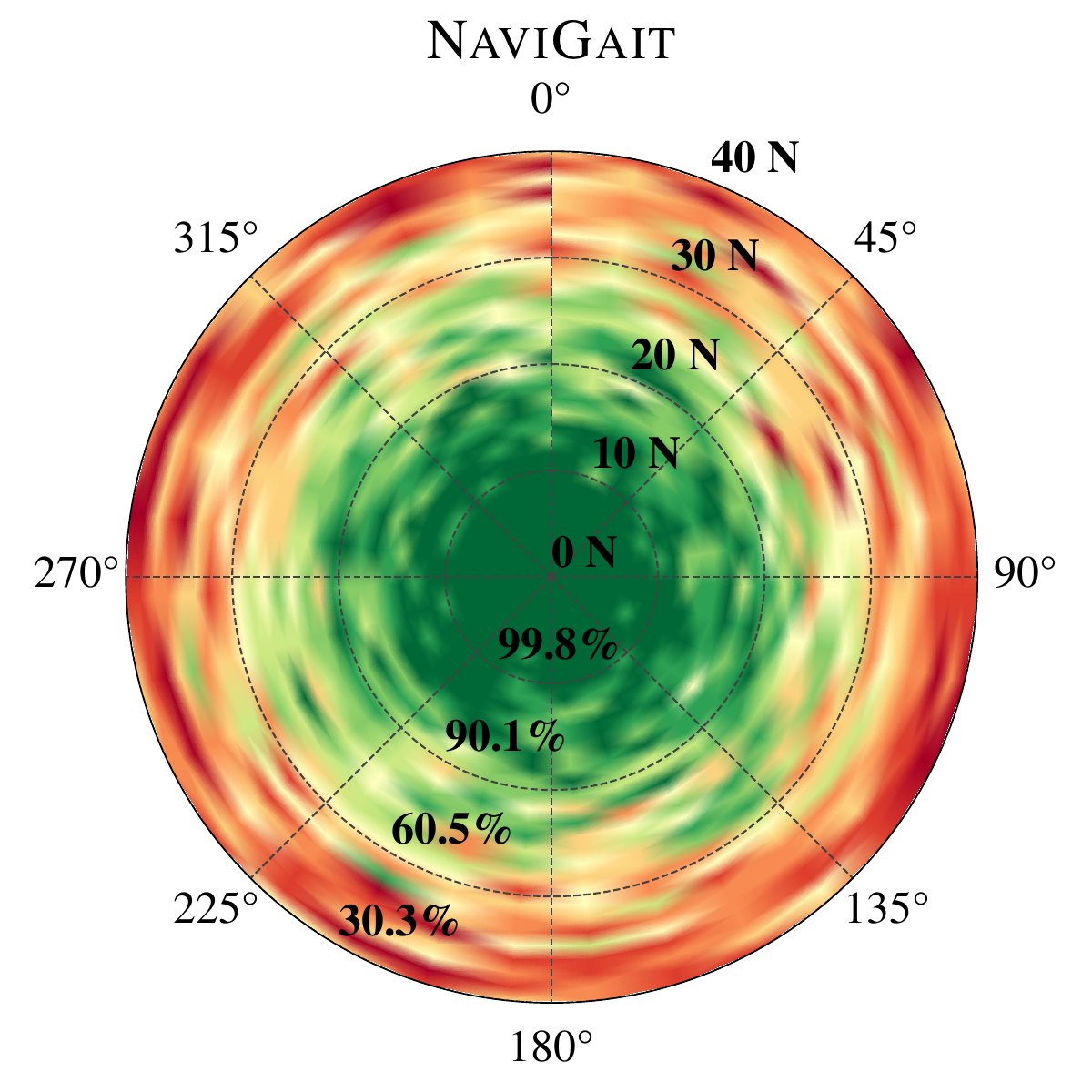}
    \end{subfigure}%
    \begin{subfigure}[htbp]{0.3\textwidth}
        \centering
        \includegraphics[width=\linewidth]{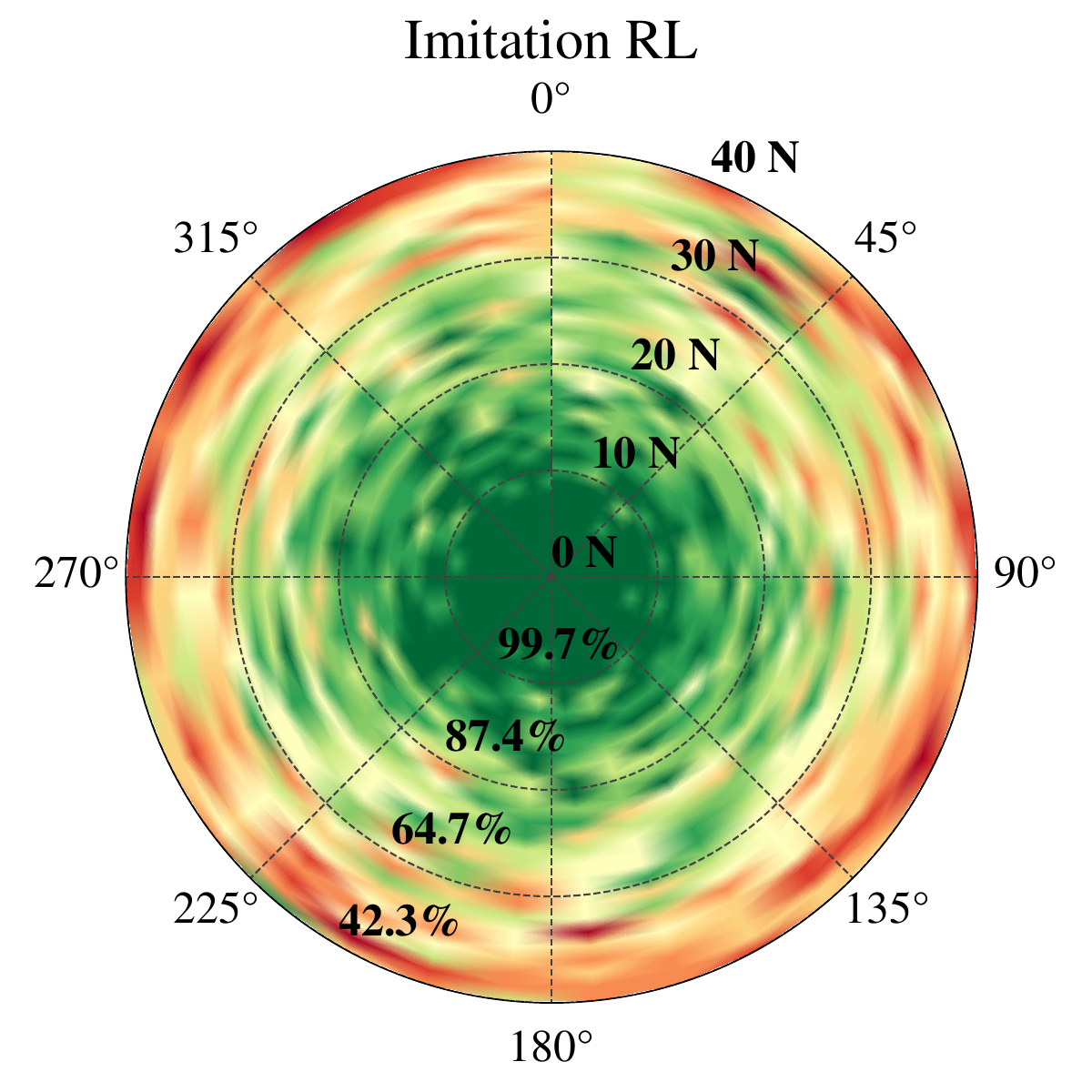}
    \end{subfigure}%
    \begin{subfigure}[htbp]{0.3\textwidth}
        \centering
        \includegraphics[width=\linewidth]{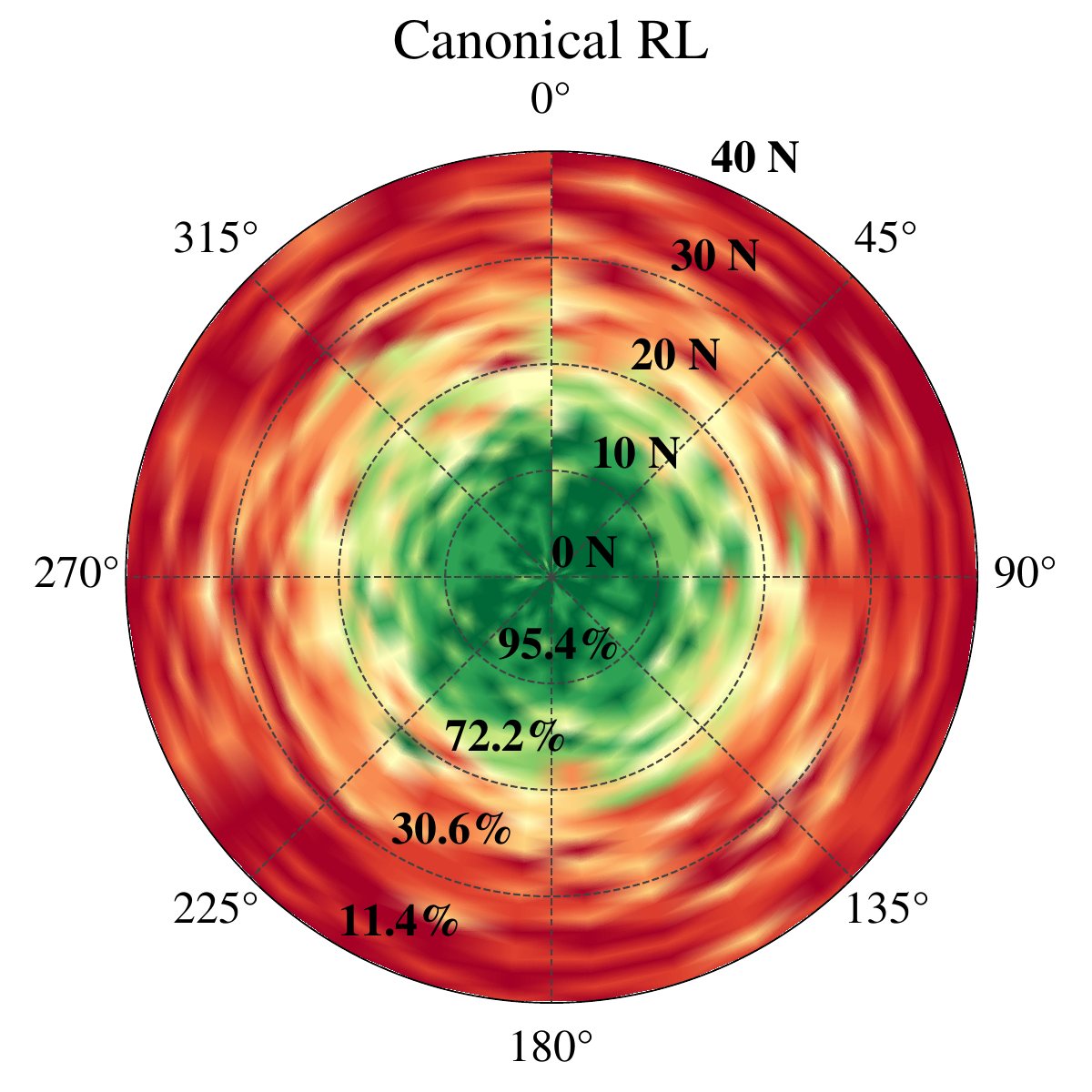}
    \end{subfigure}%
    \begin{subfigure}[htbp]{0.1\textwidth}
        \centering
        \includegraphics[width=\linewidth]{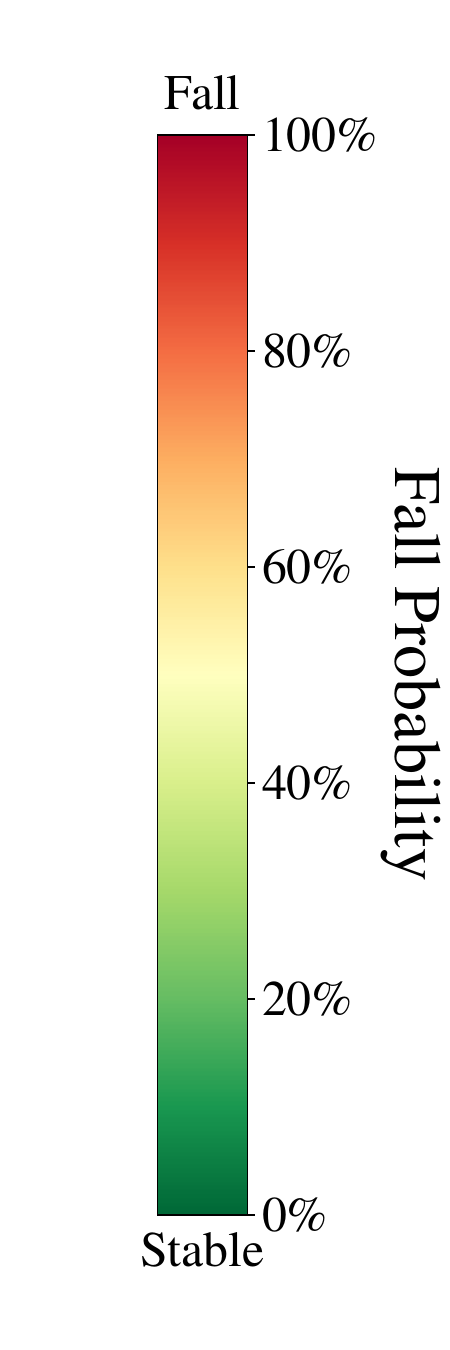}
    \end{subfigure}%
    \caption{Fall probability when applying a randomized force vector to \navigait~(left), Imitation RL (middle), and Canonical RL policy (right). The radial axis shows the magnitude of force and percentage of trials that successfully continued walking after perturbation. The \navigait~policy and Imitation RL policy both achieve similar levels of robustness to disturbances, and outperform the Canonical RL. }
    \label{fig:perturbation}
    \vspace{-4mm}
\end{figure*}


\section{Experimental Results}
\label{sec:result}

Our experiments demonstrate the performance, learning efficiency, and naturalness of the \navigait~system.
We compare the \navigait~system to two other bipedal locomotion approaches: one with no gait references (as in \cite{zakka2025mujoco}) and one with gait references provided as imitation targets (as in \cite{li2021reinforcement}).
We will refer to the first as ``Canonical RL" and the second as ``Imitation RL."  For repeatability, all comparisons are performed in a MuJoCo simulation environment. \ifbool{codelinks}{Moreover, we include all learning, domain randomization, and reward parameters in configuration files within \navigait~repository$^{(\ref{navigait-code})}$ to reproduce our results.}{}

Lastly, we transfer the \navigait~policy onto the BRUCE hardware.
We successfully demonstrate stable stepping and disturbance rejection, shown in Fig. \ref{fig:hero}.
Additional hardware demonstrations can be seen in the supplementary video.

\subsection{Baseline Setup}
For Imitation RL, we base our implementation off of \cite{li2021reinforcement} and change the residual component to directly output desired motor positions.
The gait library component is retained from \navigait.
Our reward structure is as follows.

\small
\begin{align}
    R_{\textrm{Imitation}}^t(s, a) =~&R_{\text{Gait}}^t + R_{\text{Base Cartesian}}^t + R_{\text{Base Orientation}}^t + \\
    &R_{\text{torque}} + R_{\Delta q \text{ rate}}. \notag 
\end{align}
\normalsize

For Canonical RL, we base our reward structure off \cite{zakka2025mujoco} with the following rewards.
The first four terms are implemented exactly as in \cite{zakka2025mujoco}.
The last two reward terms promote periodic stepping and reward the feet for remaining parallel to the ground.

\small
\begin{align}
    R_{\textrm{Canonical}}^t(s, a) =~&R_{\text{Linear Velocity}}^t + R_{\text{Angular Velocity}}^t + \\ 
    & R_{\text{Joint Imitation}}^t + R_{\text{action rate}} + R_{\text{stepping}} + R_{\text{feet flat}} \notag
\end{align}
\normalsize

\subsection{Walking Performance}
Our first indicator of walking performance is base velocity command tracking.
Specifically, we give all 3 policies an ovular velocity command (sinusoids in the sagittal and frontal planes) and record the actual base velocities.
The result is illustrated in Fig. \ref{fig:velocity-tracking}.
We observe that, as expected, the policies all have similar capabilities in velocity tracking, but \navigait~and Imitation RL have considerably less drift. 
This is likely due to the Canonical RL policy's tendency to maximize the velocity tracking reward for part of each step, and ignore it for the rest of the period.
This leads to the Canonical RL policy not matching the commanded velocity on average.
The reference-based methods do not suffer from this problem because the offline generated references are constrained to match the desired average walking velocity.

Our second indicator of walking performance is disturbance rejection. Over 8192 parallel trials, we command the robot to track a random velocity in a domain randomized environment and apply a force with uniform random magnitude and direction to the robot at a random time.
From the heatmap of fall probability in Fig. \ref{fig:perturbation}, we observe that \navigait~has similar overall disturbance rejection capability to the Imitation RL baseline. \navigait~demonstrates superior robustness for moderate pushes--which require corrective motions that can be sourced from the reference library--and Imitation RL is more robust against extreme pushes.
Prior work has highlighted this limitation--that residual control limits robustness compared to non-residual control since learned motions are constrained to be near the reference \cite{li2021reinforcement}. Our results show that \navigait~alleviates this limitation, shown by the increased robustness for moderate disturbances. This phenomenon is still a fundamental tradeoff, however, as discussed later in Sec \ref{sec:limitations}.



\subsection{Comparison of Training Efficiency}
We compare the ``learning speed" of each approach by measuring the number of training iterations necessary to reach a particular proportion of the maximum possible reward.
This is possible because all positive reward terms used in the training are capped and negative terms are minimal and only relate to smoothing, leading to a realistic upper bound on the reward for each environment.

\begin{figure}[tb]
    \centering
    \includegraphics[width=\linewidth]{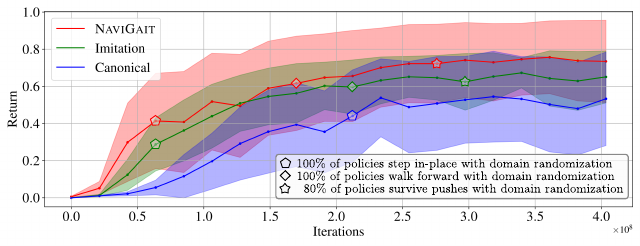}
    \caption{Comparison of the normalized
    reward return during training. At each iteration, 128 policies are evaluated under domain randomization. Markers show when 100\% of policies step in-place, 100\% walk forward, and 80\% reject perturbations. We find that \navigait~reaches these milestones earlier than Imitation RL, while Canonical RL only achieves the stepping benchmark.
    \label{fig:training-speed}} 
    \vspace{-4mm}
\end{figure}

Fig. \ref{fig:training-speed} shows the average reward per training iteration, normalized from 0 to the maximum reward.
As expected, both the \navigait~and Imitation RL policy can increase cumulative reward quickly because the reward structure more efficiently guides policy exploration towards feasible walking gaits.
Importantly, the \navigait~policy learns fundamental locomotion skills, such as stepping in-place, walking forward, and perturbation rejection, faster than both the Imitation RL and Canonical RL policies. 
Despite requiring less training time, \navigait~resulted in more natural looking walking motions as illustrated in the supplemental video.

For fair comparison, we also consider wall-clock time to stable stepping behavior: 23 minutes for \navigait, 22 minutes for Imitation RL, and 55 minutes for Canonical RL.
All policies were trained on a desktop computer with AMD Ryzen Threadripper Pro 5945WX 4.1~GHz CPU, and NVIDIA RTX A4000 GPU.
Additionally, each gait takes approximately 0.65 minutes to generate, with 85 gaits in the gait library. 
Notably, the gait generation could be significantly accelerated by generating optimized C++ code for the dynamics and constraints using C-FROST \cite{hereid2019rapid}.



\subsection{Stylistic Properties}
\begin{figure}[tb]
\vspace{1.2mm}
    \centering
    \includegraphics[width=\linewidth]{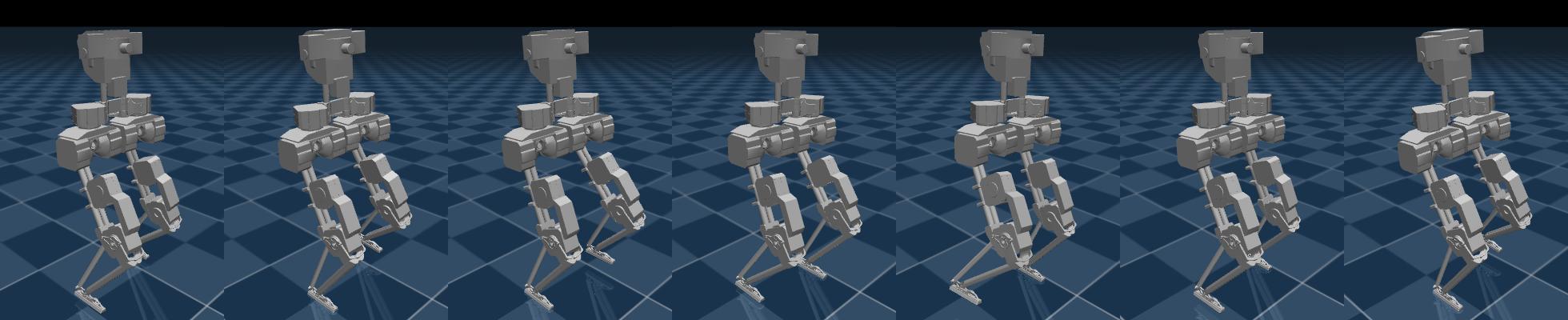}
    \includegraphics[width=\linewidth]{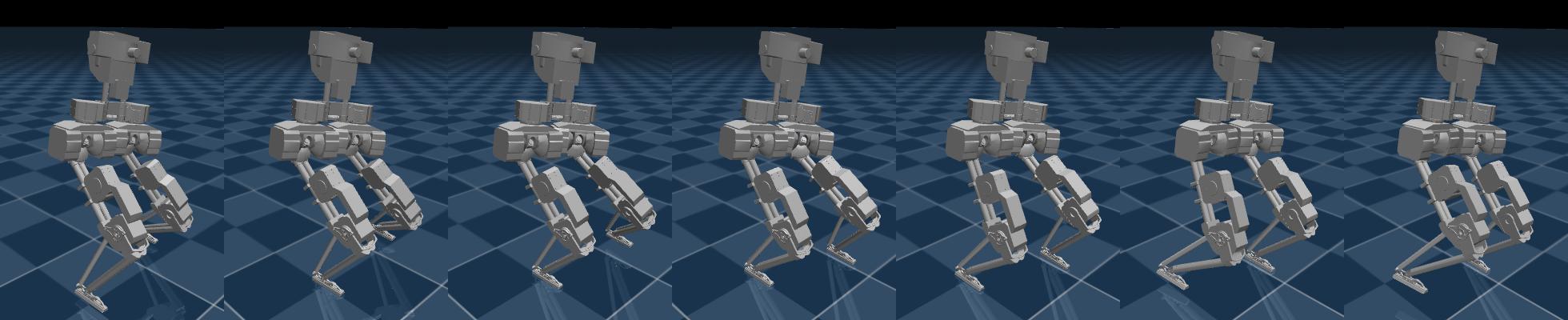}
    \caption{Visualization of two different gait libraries with different stylistic features. The top features a more ``natural" style, while the bottom has exaggerated hip roll. \navigait~makes it easy to tune these gait styles by changing costs and constraints in the NLP. }
    \label{fig:style_gaittiles}
    \vspace{-2mm}
\end{figure}

Lastly, we perform two additional experiments to demonstrate our approach's benefits in generating gaits with desirable stylistic properties.
First, because our architecture decouples the gait generation from the stabilization, we can get policies with different gait properties by simply replacing the gait library and retraining (i.e., without having to adjust weights). 
To demonstrate this, we generate two gait libraries with different stylistic differences and train an instance of \navigait~for each without changing the reward weights.
A visualization of the resulting gaits is presented in Fig. \ref{fig:style_gaittiles}.
Both policies are stable but exhibit significant stylistic differences.
This capability permits much quicker iteration of new stylized walking policies.

\begin{figure}[tb]
    \centering
    \begin{subfigure}[htbp]{0.45\textwidth}
        \centering
        \includegraphics[width=\linewidth]{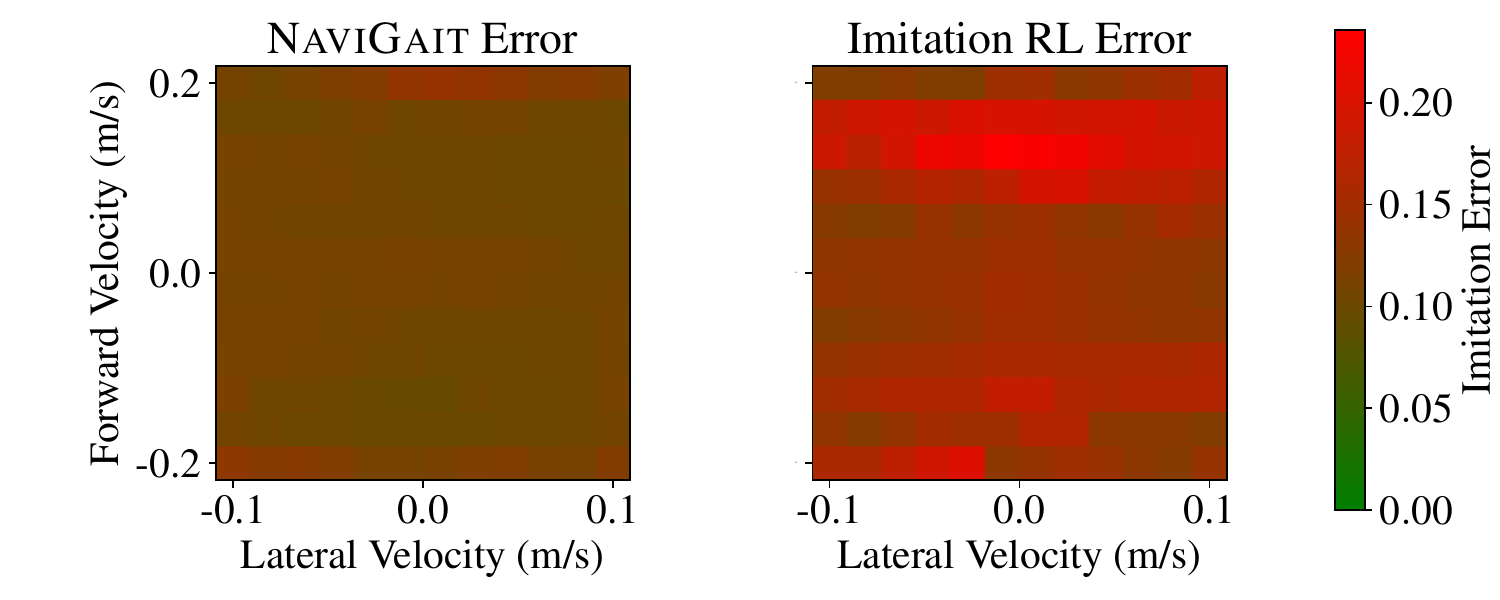}
    \end{subfigure}%
    \hfill
    \begin{subfigure}[htbp]{\linewidth}
        \centering
        \includegraphics[width=\linewidth]{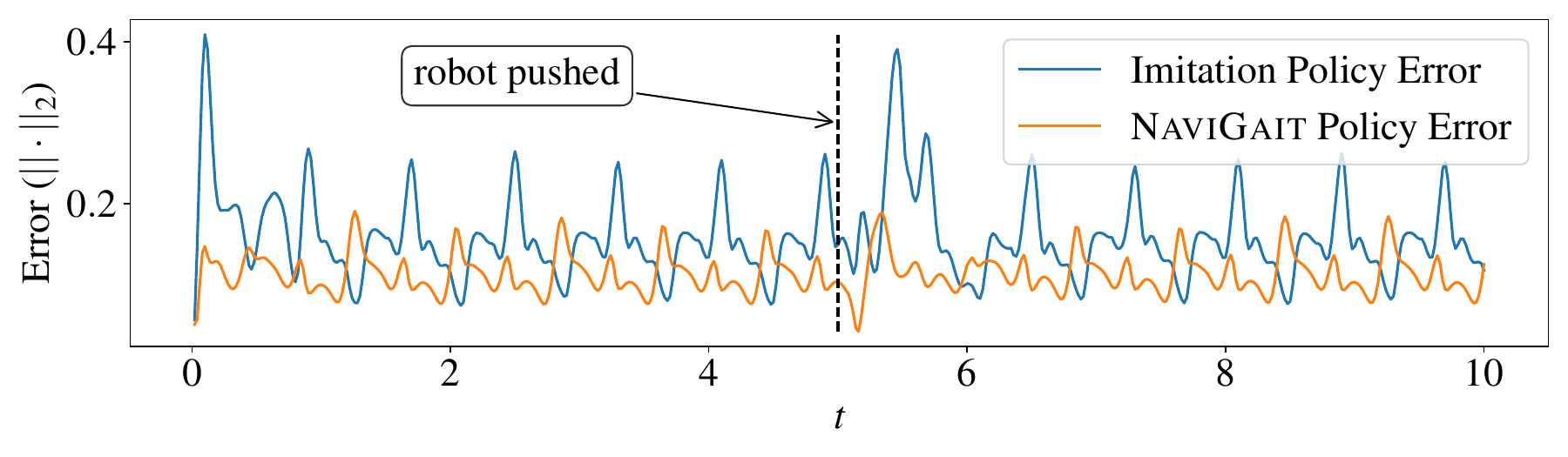}
    \end{subfigure}%
    \caption{Imitation error analysis results for the imitation-based policy and the \navigait~policy. The \navigait~policy achieves greater imitation accuracy compared to the Imitation RL approach across the gait library and during pushes, indicating it remains closer to the original reference motions. }
    \label{fig:imitation}
\end{figure}

Secondly, we show that the style is preserved even during disturbance rejection by measuring imitation accuracy, the error between the achieved and reference gait, when the robot is pushed. Fig. \ref{fig:imitation} demonstrates that \navigait~maintains a low imitation error compared to Imitation RL after the robot is pushed at $t=5$ since \navigait~stabilizes by shifting to a different gait in the gait library.
The heatmaps in \ref{fig:imitation} also demonstrate that \navigait~consistently outperforms Imitation RL in tracking ability across the entire gait library, leading to better expression of the gait library style.


\section{Limitations}
\label{sec:limitations}
The primary limitation of \navigait~is that our architecture limits the capability of the RL agent to learn emergent behaviors such as foot cross-over.
The generated gait library does not contain any ``cross-over'' stepping behaviors and the training includes a penalty which encourages the output trajectories to remain close to those in the gait library.
This may limit the robustness of the controller.
We see this as a fundamental tradeoff between predictability and preservation of gait style and the emergence of novel behaviors.
In some situations, like wearable robots with human subjects, we prefer to have policies that are predictable.
Additionally, setting up and tuning the trajectory optimization problem requires additional domain expertise.
To combat this, we have provided our gait generation framework for reference. 

\section{Conclusion}
\label{sec:conclusion}
    
In this work, we present \navigait, a locomotion framework that combines residual reinforcement learning with a precomputed gait library to enable reliable and natural-looking walking on the BRUCE Humanoid robot.
Motivated by the architecture of gait library regulators, our approach leverages offline gaits which are easy to tune and personalize for desired stylistic changes.
However, unlike gait library regulators which rely on heuristic corrections, \navigait~leverages RL to improve robustness and adaptability.
Our choice of architecture decouples the behavior shaping aspect from the stabilization aspect of the policy, resulting in policies that are grounded in interpretable trajectory optimization and thus can be easily tuned, while still affording robustness.
Overall, this architecture preserves style while generating corrective actions which were not encoded directly into the gait library.

We experimentally demonstrate that \navigait~achieves better performance faster during training, comparable robustness to conventional reinforcement learning approaches, and improved gait naturalness, despite using a simpler reward structure.
The developed framework provides a foundation for future extensions, including expressive and personalized gaits that are relevant in character animation, animatronics, and wearable robots.
Overall, \navigait~offers a promising direction for locomotion by bridging the gap between handcrafted motion planning and end-to-end learning.

\section*{ACKNOWLEDGMENT}

\ifanonymous
\else
We thank Gary (Lizhi) Yang, Morgan Byrd, and Yusuke Tanaka for their helpful discussions on sim-to-real transfer.
\fi

\bibliographystyle{ieeetr}
\bibliography{example}  

\end{document}